\documentclass[]{ceurart}

\usepackage{listings}
\begin{document}

\copyrightyear{2026}
\copyrightclause{Copyright for this paper by its authors.
	Use permitted under Creative Commons License Attribution 4.0
	International (CC BY 4.0).}

\conference{SuRE'26: Workshop on Sustainability and Resource-Efficiency of Artificial Intelligence, August 17, 2026, Bremen, Germany}

\title{Accounting for Bias Enables Sustainable LLM Evaluation}


\author[1,2]{Harshita Katoch}[%
  email=harshita.katoch@dfki.de,
]
\author[1,2]{David Antony Selby}[%
  email=david.selby@dfki.de,
]
\author[2]{Gerrit Gro{\ss}mann}[%
  email=gerrit.grossmann@dfki.de,
]
\author[1,2]{Sebastian Vollmer}[%
  email=sebastian.vollmer@dfki.de,
]
 
\address[1]{Department of Computer Science,
  University of Kaiserslautern--Landau (RPTU), Germany}
\address[2]{Data Science and its Applications,
  German Research Centre for Artificial Intelligence (DFKI), Germany}


\begin{abstract}
	LLM-as-a-judge has become the \textit{de facto} standard for scalable, subjective evaluation, yet current leaderboards compensate for systematic measurement bias by running ever more comparisons, an approach that is both statistically unsound and computationally wasteful. The root cause is an incomplete measurement model, treating LLM judges as neutral, interchangeable instruments ignores documented biases like position bias, verbosity bias, judge severity, and self-enhancement, that no volume of additional data can eliminate. We propose a unified latent variable framework that jointly models pairwise and ordinal data while explicitly correcting for these confounders, recovering reliable rankings from substantially fewer comparisons. Because fitting this model costs negligible compute relative to a single round of LLM inference, bias correction is not only more statistically rigorous but also a more sustainable approach to trustworthy evaluation.
\end{abstract}


\maketitle

\section{Introduction}

The evaluation of large language models has become a question of trustworthiness as much as capability~\cite{ye:llm-psychometrics}. The field's prevailing response to evaluation challenges has been to scale: running more comparisons, enlisting more judges and expanding leaderboards. The statistical machinery underpinning these systems has genuinely 
improved, with the field moving from raw win rates toward principled latent-variable models for pairwise and ordinal scoring that better capture preference uncertainty. Yet scaling and methodological refinement have both proceeded under a shared assumption, that has gone largely unexamined: that LLM judges are neutral, interchangeable instruments whose idiosyncrasies cancel with sufficient data~\cite{gu:survey,li:llms-as-judges}. When judge behaviour is systematically skewed, collecting more comparisons narrows the confidence interval around a biased estimate but not correct it. Scale adds compute without adding reliability.

The biases in question are well-documented and structural. LLM judges exhibit position preferences i.e. favouring responses that appear in a particular slot regardless of quality~\cite{shi:position-bias}, verbosity effects that reward length over content~\cite{dubois:alpacaeval}, and self-enhancement toward outputs from their own provider family~\cite{marsh:ikea}. Beyond these, different judges apply systematically different score levels which is a severity effect that makes identical model ability map to different observed scores. Surveys by \citeauthor{gu:survey}~\cite{gu:survey} and \citeauthor{li:llms-as-judges}~\cite{li:llms-as-judges} establish these as properties of the measurement instrument, not incidental quirks of particular models. Since the effects differ in direction across judges, procedural fixes such as order randomisation or majority voting cannot eliminate them in expectation~\cite{li:llms-as-judges}, nor do mitigations that target a single bias in isolation~\cite{dubois:alpacaeval,boyeau:autoeval}. A jointly specified measurement model is required.

The consequences are both scientific and environmental. Every judge-mediated comparison requires inference from two competing models and a judge, i.e. three inference calls per comparison, and at leaderboard scale this cost is substantial.
Each inference call consumes computational resources, GPU time, memory, storage and electricity (training costs notwithstanding), so reducing the number of comparisons translates directly into lower energy use and a smaller environmental footprint~\cite{samsi2:words}.
Standard benchmarks are approximately 97\% redundant~\cite{kipnis:metabench}, and smarter item selection can reduce evaluation cost by up to 99\% without accuracy loss~\cite{polo:tinybenchmarks}. Generating redundant comparisons to compensate for an unmodelled confounder is not just statistically unsound, it is ecologically indefensible~\cite{bender:parrots,burnell:rethink-reporting}. Fitting a corrective latent-variable model costs negligible compute relative to a single round of LLM inference, making bias correction the more sustainable path to evaluation we can actually trust.

We propose a unified latent-variable framework that (i)~jointly models pairwise and ordinal data in a single likelihood, (ii)~explicitly corrects for position bias, verbosity, self-enhancement, and judge-level severity, and (iii)~is validated on MT-Bench~\cite{zheng:mt-bench} as the first empirical decomposition of bias magnitudes on a real leaderboard. Simulation confirms that the bias-corrected model recovers reliable rankings from substantially fewer comparisons than naive baselines across six tournament designs.

\section{Related Work}
The machine learning community has developed LLM evaluation largely in isolation from the statistical sciences, arriving at methods that parallel tools already mature in psychometrics and educational measurement~\cite{jordan:computational}. Early evaluation relied on static benchmarks with ground-truth answers; as tasks became more open-ended, the field shifted toward judge-mediated evaluation, MT-Bench introduced GPT-4-as-a-judge on multi-turn dialogue~\cite{zheng:mt-bench}; Chatbot Arena crowdsourced human pairwise preferences into Elo scores~\cite{chiang:chatbot-arena}; AlpacaEval~2.0 and Arena-Hard extended the paradigm with length control and harder prompts~\cite{dubois:alpacaeval,li:arena-hard}. Each iteration improved on its predecessor, yet all share a common assumption that observed ratings are direct proxies for model quality rather than noisy realisations of a latent construct mediated by rater behaviour and item difficulty. Beyond the response-level biases documented in the literature~\cite{gu:survey,li:llms-as-judges}, the measurement scale itself is routinely misused, \citeauthor{liddell:ordinal}~\cite{liddell:ordinal} shows that the widespread practice of treating ordinal judge scores as continuous interval data introduces systematic distortions into effect estimates, an error that accumulating more data cannot fix. A jointly specified model is required. 

Item response theory disentangles examinee ability from item difficulty, providing estimates invariant to the specific sample of items or judges used~\cite{demars:irt}. Many-facet Rasch models extend this to explicitly model rater severity~\cite{martinez-plumed:irt-ml}, and extensions of paired-comparison models provide a principled framework that subsumes Elo as a special case~\cite{hamilton:comparative-judgment}. Recent work has begun applying these ideas to AI benchmarks~\cite{lalor:irt-nlp,federiakin:llm-leaderboards,wang:psychometrics-ai}, but exclusively on objective tasks with binary correctness signals, where there is no rater in the loop. Our contribution extends this to the judge-mediated setting, where rater confounders require explicit parametrisation alongside standard IRT machinery, a gap the existing literature has not addressed.

\section{A Unified Psychometric Framework}

In psychometric terms, the LLM under evaluation is the \textit{examinee} with latent ability~$\theta$, the prompt is the \textit{item} with difficulty~$\delta$, and the judge is the \textit{rater} with severity~$\alpha$. The goal is \textit{parameter invariance}: ability estimates independent of which specific judges and prompts happened to be used~\cite{demars:irt}. If a model's ranking drops because we switched from a lenient judge to a strict one, the measurement system has failed. This reframes what appears to be a data-volume problem as what it actually is a modelling problem. 

\subsection{Ranking from Pairwise Comparisons}

A minimally viable evaluation system based on comparative judgement should use a variant of the Bradley--Terry model~\cite{bradley:terry} rather than average win rates, with coefficients to account for order effects, self-enhancement, and measurable item-level features such as relative verbosity. This extends AlpacaEval~2.0~\cite{dubois:alpacaeval}, which in the presence of these biases only controls for item-level effects. Such an augmented Bradley--Terry--Luce model takes the form:
\begin{equation}
P(A \succ B) = \sigma\!\left(
  \theta_A - \theta_B + \gamma +
  \underbrace{\lambda_{A,r} - \lambda_{B,r}}_{\text{Affinity bias}}
\right)
\label{eq:bt}
\end{equation}
where $\theta_A$ and $\theta_B$ are the models' latent abilities;
$\gamma$ is a position bias; $\lambda_{i,r}$ is judge~$r$'s affinity for model~$i$, capturing self-enhancement bias; and $\sigma$ is the logistic function. This is straightforward to fit using standard logistic regression software. Further terms may be added for verbosity and other documented biases; in our empirical analysis we additionally include a judge-specific position deviation~$u_j$ and a standardised response length covariate~\cite{dubois:alpacaeval}.

\subsection{Ranking from Ordinal Scores}

Some benchmarks prefer direct scoring of individual items. A comparative judgement may say $A \succ B$ but not be able to say `both are bad'; an absolute scale such as 1--10 allows us to anchor items at one or the other extreme rather than merely order them relative to one another. Instead of naively averaging scores, a minimal ordinal model must account for scale-, item-, and judge-level effects. The many-facet Rasch model~\cite{linacre:mfrm} treats scores as ordered categorical outcomes. The probability that model~$n$ receives score $k \in \{0, \dots, m\}$ from rater~$r$ on item~$p$ is:
\begin{equation}
P(X_{npr} = k) \propto
  \exp\!\left(
    \sum_{m=1}^{k}
    (\theta_n - \delta_p - \alpha_r - \tau_m)
  \right)
\label{eq:mfrm}
\end{equation}
where $\theta_n$ is latent ability, $\delta_p$ is item difficulty, $\alpha_r$ is rater severity, and $\tau_m$ are threshold parameters on the cumulative logit scale, not assumed equally spaced. Accounting for judge severity prevents a model evaluated by a lenient judge from artificially outranking one evaluated by a strict judge. Treating ordinal scores as interval data, as leaderboards routinely do, produces systematic inversions in estimated effects~\cite{liddell:ordinal}, a modelling error that accumulating more data cannot fix.

\subsection{The Hybrid Joint Likelihood}

Both pairwise preferences and ordinal scores are imperfect observations of the same latent ability. Rather than discarding either, we combine them in a joint likelihood that separates true ability variance from rater noise:
\begin{equation}
\mathcal{L}(\theta) =
  \!\left[
    \prod_{y \in \mathcal{D}_\mathrm{pair}}
    \!P_\mathrm{Rel}(y \mid \theta, \gamma)
  \right]^{\omega}
  \!\times\!
  \left[
    \prod_{x \in \mathcal{D}_\mathrm{ord}}
    \!P_\mathrm{Ord}(x \mid \theta, \lambda, \delta)
  \right]^{1-\omega}
\label{eq:hybrid}
\end{equation}
where $\omega \in [0,1]$ weights the two data sources ($\omega = 0.5$ in our experiments); $\lambda$ represents judge-specific parameters (severity and affinity); and $\delta$ item difficulties. Ordinal ratings anchor sparse pairwise data, and models appearing in only one data source can be ranked alongside those in both. Estimation proceeds via L-BFGS-B with ridge regularisation; for adaptive designs, a parametric bootstrap correction is preferred over Firth's penalised likelihood, which introduces bias under non-random scheduling~\cite{hamilton:comparative-judgment}.

\section{Empirical Validation: MT-Bench}
\label{sec:empirical}

We validate the proposed framework on MT-Bench~\cite{zheng:mt-bench}, fitting the hybrid model jointly to GPT-4-judged pairwise comparison data and single-score ordinal data. Our aim is to confirm that the documented judge confounders are statistically significant in real leaderboard data, that they produce systematic, model-specific distortions rather than random noise, and that correcting for them recovers meaningful signal that no amount of additional data collected under the same protocol could provide. 

\paragraph{Judge confounders are structural, not stochastic.} Table~\ref{tab:biasparams} reports the estimated parameters from the full model. Position bias ($\hat{\gamma}_0 = -0.165$) confirms that second-position responses are systematically favoured, consistent with the judge-dependent direction effects documented by \citeauthor{shi:position-bias}~\cite{shi:position-bias}. Verbosity bias ($\hat{\beta} = 0.163$, 95\,\% CI\,$[0.058, 0.268]$) confirms that longer responses are preferred irrespective of content quality, extending the finding of \citeauthor{dubois:alpacaeval}~\cite{dubois:alpacaeval} to a different evaluation setting. Same-family affinity ($\hat{\lambda} = 0.411$) confirms that the GPT-4 judge systematically favours models from its own provider family, a direct violation of the measurement invariance property that any reliable leaderboard requires. Substantial judge-level variation in position sensitivity (SD$(u_j) = 0.337$) compounds this further, in the MT-Bench data, \texttt{claude-v1} appeared in second position far more often than first while \texttt{llama-13b} faced the opposite exposure, so the global position parameter accrues as a tournament-design artefact rather than a model quality signal. None of these effects can be eliminated by collecting more data under the same evaluation protocol, because they are properties of the measurement instrument rather than of sampling variation.

\begin{table}[t]
  \centering
  \caption{Estimated bias parameters from the hybrid model on MT-Bench. The parameter estimates are statistically significant or substantially non-zero, confirming that LLM-judge evaluation is not bias-free at any sample size.}  
  \label{tab:biasparams}
  \begin{tabular}{lrrr}
    \toprule
    Parameter & Estimate & \multicolumn{2}{c}{95\,\% CI} \\
    \midrule
    $\hat{\gamma}_0$ (position bias)       & $-0.165$ & $-0.253$ & $+0.076$ \\
    $\hat{\beta}$ (verbosity)              & $+0.163$ & $+0.058$ & $+0.268$ \\
    $\hat{\lambda}$ (same-family affinity) & $+0.411$ & --- & --- \\
    $\mathrm{SD}(u_j)$ across judges                & $0.337$  & --- & --- \\
    \bottomrule
  \end{tabular}
\end{table}

\paragraph{Bias correction recovers signal that more data cannot.}
\begin{figure}[t]
  \centering
  \includegraphics[width=\linewidth]{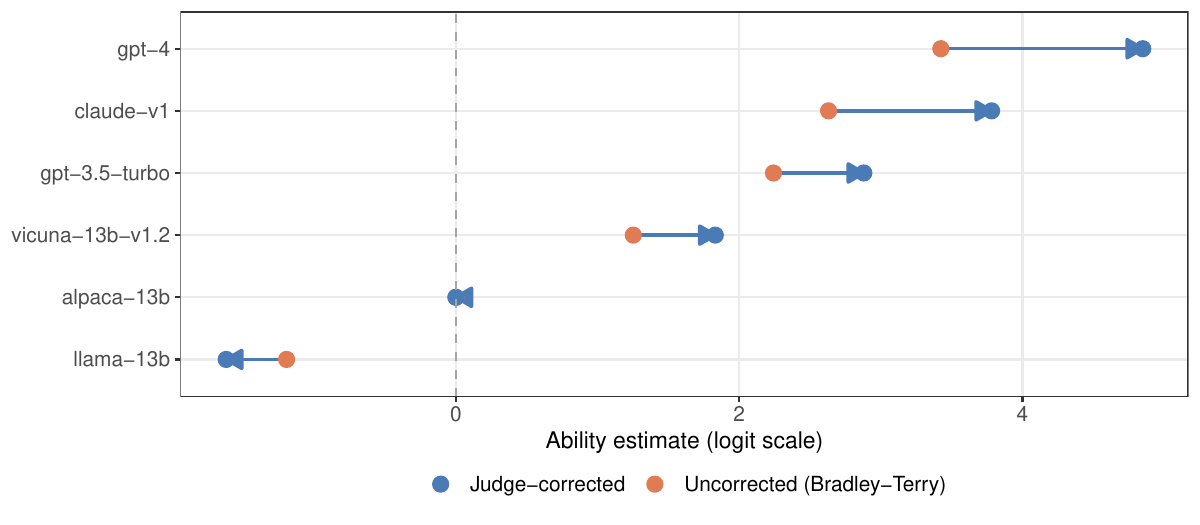}
  \caption{%
    \textbf{Correcting for judge effects shifts ability estimates substantially.} Each arrow shows the change from naive Bradley-Terry (orange) to the Judge-corrected model (blue). Provider-affiliated models receive the largest upward corrections; non-affiliated models shift downward. These distortions are structural and collecting more comparisons under the same protocol would not resolve them.}
  \label{fig:dumbbell}
\end{figure}
Figure~\ref{fig:dumbbell} shows the net outcome of jointly correcting for all three bias components. Ability estimates shift substantially and in model-specific directions that trace directly back to tournament design rather than model quality. Critically, this signal recovery requires no additional data: the bias-corrected model is fitted to the same battles as the naive baseline. The two frameworks disagree on a non-trivial share of model pairs (Kendall $\tau = 0.80$) even when all available data are used, a structural disagreement that persists regardless of how many observations each framework receives. When bias is structural, the path to reliable rankings runs through better measurement, not more measurement.

\section{Sustainability: Fewer Comparisons}
\label{sec:sustainability}

The measurement argument has a direct sustainability consequence. A misspecified model treats structural rater noise as signal, so every comparison it schedules partly compensates for bias rather than resolving genuine uncertainty about model ability. A correctly specified model never accumulates this waste in the first place, because it attributes rater noise to the judge rather than to the models being compared. The result is stable rankings from fewer comparisons and, at leaderboard scale, fewer comparisons means substantially less LLM inference.

\begin{figure*}[ht] 
  \centering
 \includegraphics[width=\linewidth]{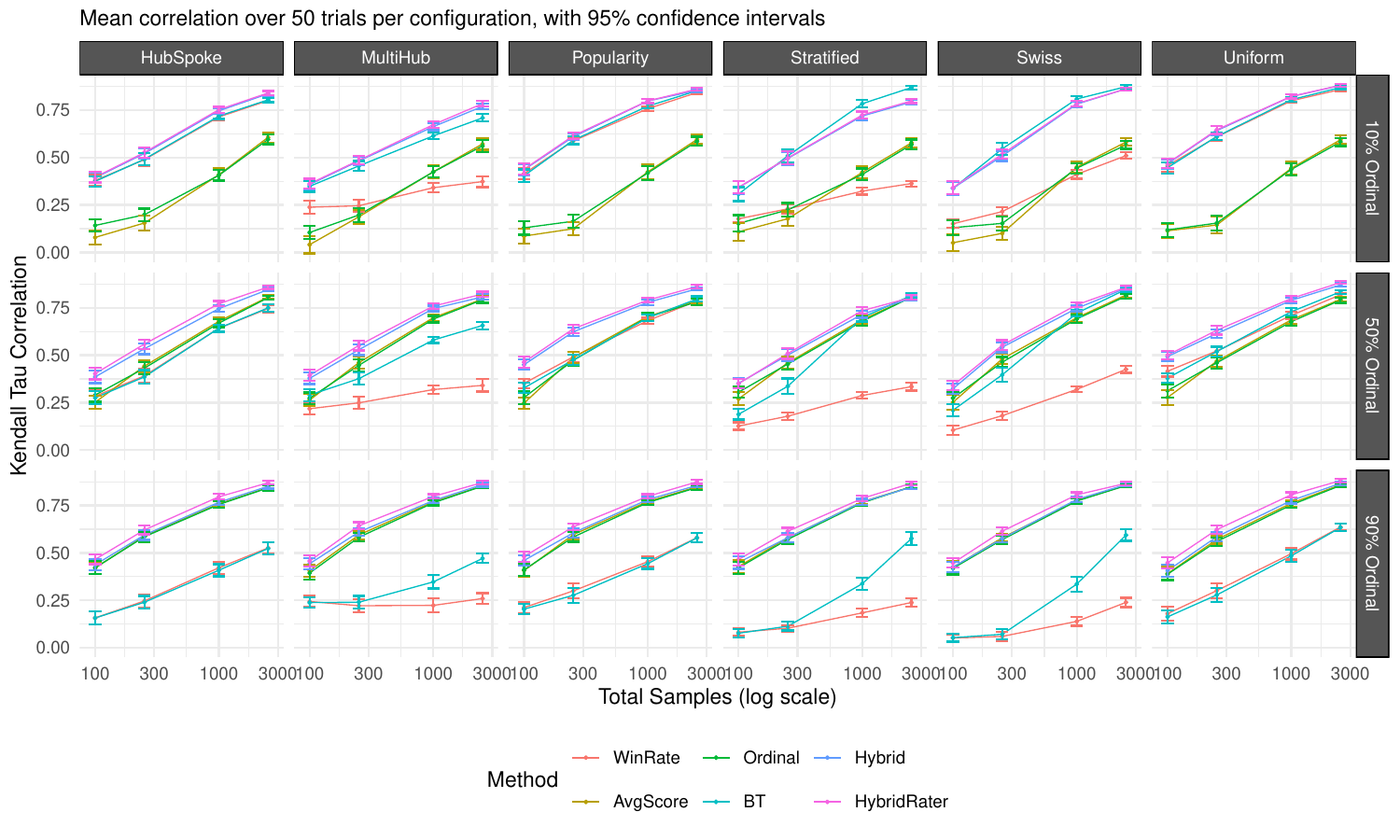}
  \caption{%
    \textbf{Hybrid model recovers ground-truth rankings with fewer comparisons.} Mean Kendall-$\tau$ correlation with ground truth over 50 trials, shown across six tournament structures and three proportions of ordinal data. HybridRater consistently outperforms naive win-rate and simple score averaging, particularly in sparse,     adversarially scheduled designs that resemble real leaderboard conditions. The efficiency gain is largest where it matters most, at low sample sizes and in unbalanced tournament designs. Careful selection of the scaling parameter~$\omega$ is important when combining the two measurement scales.}
  \label{fig:simulation}
\end{figure*}

To quantify this, we ran a simulation of $N = 30$ models across six tournament designs (Uniform, Swiss, Stratified, HubSpoke, MultiHub, Popularity) and three mixes of ordinal and pairwise data (10\%, 50\%, and 90\% ordinal), averaging Kendall-$\tau$ correlation with ground truth over 50 trials per configuration. Figure~\ref{fig:simulation} reports the results. Across almost all settings, the bias-corrected hybrid model (HybridRater) recovers stable rankings at sample sizes where naive win-rate and simple score averaging remain well below equivalent fidelity. The gap is largest in the adversarial designs- Swiss, Stratified, and MultiHub which most closely resemble real leaderboard conditions where scheduling is non-uniform and model exposure is unbalanced. In these regimes, naive aggregation plateaus well below the hybrid model even at the largest sample sizes tested, confirming that the efficiency gain is not merely a matter of getting there faster, it is a matter of getting there at all.

Real leaderboards can recover further gains through adaptive scheduling, rather than drawing pairs at random, new comparisons can be targeted where uncertainty in~$\theta$ is highest, and the Fisher information of the joint latent-variable model provides a principled criterion for this. The connection to computerised adaptive testing in educational measurement~\cite{wainer:cat} is direct. \citeauthor{zhuang:ai-eval-humans}~\cite{zhuang:ai-eval-humans} argue that AI evaluation should adopt precisely these methods from human psychometric testing; \citeauthor{polo:tinybenchmarks}~\cite{polo:tinybenchmarks} demonstrate their practical value, achieving 99\% cost reduction through IRT-based item selection on static benchmarks; and \citeauthor{li:active-eval}~\cite{li:active-eval} extend this to a fully active setting, framing evaluation as a sequential acquisition problem where each comparison is selected to maximise expected information gain about model ability. Chatbot Arena already exploits a heuristic version of this, reporting a 54\% reduction in required battles through adaptive sampling~\cite{chiang:chatbot-arena}; the psychometric framework proposed here replaces that heuristic with a grounded information-theoretic criterion. Full implementation of adaptive scheduling within the hybrid framework is an important direction for future work.

\section{Limitations and Open Problems}

\paragraph{Unidimensionality.}
The framework assumes a single latent ability dimension~$\theta$. For models evaluated across diverse tasks, a multidimensional IRT model would be more appropriate: \citeauthor{burnell:capabilities}~\cite{burnell:capabilities} show via factor analysis that LLM capabilities have non-trivial latent structure, making a single~$\theta$ an approximation. Extending the hybrid framework to MIRT is an important open direction. 
\paragraph{Construct validity.}
Correcting for known confounders is necessary but not sufficient. Whether the corrected~$\theta$ tracks the construct of interest requires systematic validation not yet done at scale for LLM-as-a-judge evaluation~\cite{wang:psychometrics-ai}. The psychometric framing at least renders this an explicit, testable property of the measurement system rather than an unexamined assumption.

\section{Conclusion}
\label{sec:conclusion}

Current leaderboards treat instability as a data problem and respond by scaling comparisons. We have argued it is a measurement problem, the biases that distort rankings are structural properties of the evaluation instrument, not sampling artefacts, and no volume of additional battles under the same protocol will correct them. On MT-Bench, position preference, verbosity, self-enhancement, and judge-level heterogeneity are simultaneously detectable, significant, and large enough to move ability estimates in directions that trace back to tournament design rather than model quality. 

The remedy is inexpensive. Fitting a generalised linear mixed model costs negligible compute relative to a single round of LLM inference, yet it separates ability from evaluator artefact in a way that arbitrarily large naive samples cannot. The unified framework proposed here---augmented Bradley--Terry models for pairwise data, many-facet Rasch models for ordinal data, and a joint likelihood that bridges both---provides the measurement foundation that current leaderboards lack. More reliable rankings, from fewer comparisons, at lower computational cost, is what sustainable evaluation looks like in practice.

\begin{acknowledgments}
	We thank the anonymous reviewers. This work was funded by the German Federal Ministry for Research, Technology and Space Travel (BMFTR) grant number \texttt{01IW23005}.
\end{acknowledgments}

\section*{Declaration on Generative AI}
During the preparation of this work the author(s) used Claude for grammar and spelling checks and minor phrasing improvements. After using the tool, the author(s) reviewed and edited the content as needed and take(s) full responsibility for the publication's content.  

\bibliography{sample-ceur}




\end{document}